\documentclass[letterpaper, 10 pt, conference]{ieeeconf}
\IEEEoverridecommandlockouts
\let\labelindent\relax
\usepackage{graphicx} \graphicspath{ {figures/} }
\usepackage{amsmath,amssymb,mathabx}
\usepackage[linesnumbered,ruled,vlined]{algorithm2e}
\usepackage{acronym}
\usepackage{enumitem}
\usepackage{booktabs}
\usepackage{hyperref}
\usepackage{balance}
\usepackage{xspace,setspace}
\usepackage[skip=3pt,font=small]{subcaption}
\usepackage[skip=3pt,font=small]{caption}
\usepackage[dvipsnames]{xcolor}
\usepackage[capitalise]{cleveref}
\usepackage{tabularx,colortbl,multirow,array,makecell}
\usepackage{overpic}
\usepackage{cite}
\usepackage{tikz}
\usepackage[]{mdframed}
\usepackage{soul}
\usepackage{anyfontsize}

\newcommand*\circled[1]{\tikz[baseline=(char.base)]{
            \node[shape=circle,draw,inner sep=0.6pt] (char) {#1};}}

\makeatletter
\DeclareRobustCommand\onedot{\futurelet\@let@token\@onedot}
\def\@onedot{\ifx\@let@token.\else.\null\fi\xspace}
\def\eg{\emph{e.g}\onedot} 
\def\ie{\emph{i.e}\onedot}

\makeatother

\crefname{algocf}{alg.}{algs.}
\Crefname{algocf}{Algorithm}{Algorithms}

\makeatletter
\def\BState{\State\hskip-\ALG@thistlm}
\makeatother

\makeatletter
\renewcommand{\paragraph}{%
  \@startsection{paragraph}{4}%
  {\z@}{0ex \@plus 0ex \@minus 0ex}{-1em}%
  {\hskip\parindent\normalfont\normalsize\bfseries}%
}
\makeatother

\definecolor{gblue}{HTML}{4285F4}
\definecolor{gred}{HTML}{DB4437}

\definecolor{custorange}{RGB}{255, 147, 30}
\definecolor{custblue}{RGB}{63, 167, 243}
\definecolor{custdarkblue}{RGB}{38, 99, 145}
\definecolor{custgrey}{RGB}{202, 202, 202}
\definecolor{custgreen}{RGB}{34, 139, 34}

\colorlet{lightorange}{custorange!20}
\newcommand{\hllo}[1]{%
    {%
    \sethlcolor{lightorange}%
    \hl{#1}%
    }%
}
\colorlet{lightblue}{custblue!20}
\newcommand{\hllb}[1]{%
    {%
    \sethlcolor{lightblue}%
    \hl{#1}%
    }%
}
\colorlet{lightgrey}{custgrey!40}

\newcommand{\framedtext}[1]{%
\par\vspace{1mm} 
\noindent\fbox{%
    \parbox{\dimexpr\linewidth-2\fboxsep-2\fboxrule}{#1}%
}%
\par\vspace{2mm} 
}

\acrodef{dof}[DoF]{Degree of Freedom}
\acrodef{vkc}[VKC]{Virtual Kinematic Chain}
\acrodef{tamp}[TAMP]{Task and Motion Planning}
\acrodef{pddl}[PDDL]{Planning Domain Definition Language}
\acrodef{rrt}[RRT]{Rapidly-exploring Random Tree}
\acrodef{ompl}[OMPL]{Open Motion Planning Library}
\acrodef{iws}[IWS]{Iterated Width Search}
\acrodef{bfs}[BFS]{Breadth First Search}
\acrodef{ai}[AI]{Artificial Intelligence}
\acrodef{llm}[LLM]{Large Language Model}
\acrodef{cot}[CoT]{Chain-of-Thought}
\acrodef{rrt}[RRT]{Rapidly-exploring Random Trees}

\let\oldnl\nl
\newcommand{\nonl}{\renewcommand{\nl}{\let\nl\oldnl}}
\newcolumntype{x}{>{\columncolor{MistyRose}}c}
\newcolumntype{y}{>{\columncolor{LightCyan1}}c}

\newcommand{\llmmm}{LLM\textsuperscript{3}\xspace}
\title{\LARGE \bf
\llmmm: Large Language Model-based Task and Motion Planning \\ with Motion Failure Reasoning}

\author{Shu Wang$^{1\star}$, Muzhi Han$^{1\star}$, Ziyuan Jiao$^{2\star^\dagger}$, Zeyu Zhang$^{2}$, Ying Nian Wu$^{1}$, Song-Chun Zhu$^{2}$, Hangxin Liu$^{2^\dagger}$ 
\thanks{$^{\star}$ Shu Wang, Muzhi Han, and Ziyuan Jiao contributed equally to this work. $^\dagger$ Corresponding authors. $^{1}$ University of California, Los Angeles.\quad $^{2}$ State Key Laboratory of General Artificial Intelligence, BIGAI.
}%
}

\begin{document}
\maketitle


\begin{abstract}
Conventional \ac{tamp} approaches rely on manually designed interfaces connecting symbolic task planning with continuous motion generation. These domain-specific and labor-intensive modules are limited in addressing emerging tasks in real-world settings.
Here, we present \llmmm, a novel \ac{llm}-based \ac{tamp} framework featuring a domain-independent interface. Specifically, we leverage the powerful reasoning and planning capabilities of pre-trained \acp{llm} to propose symbolic action sequences and select continuous action parameters for motion planning. Crucially, \llmmm incorporates motion planning feedback through prompting, allowing the \ac{llm} to iteratively refine its proposals by reasoning about motion failure. Consequently, \llmmm interfaces between task planning and motion planning, alleviating the intricate design process of handling domain-specific messages between them.
Through a series of simulations in a box-packing domain, we quantitatively demonstrate the effectiveness of \llmmm in solving \ac{tamp} problems and the efficiency in selecting action parameters. Ablation studies underscore the significant contribution of motion failure reasoning to the success of \llmmm. Furthermore, we conduct qualitative experiments on a physical manipulator, demonstrating the practical applicability of our approach in real-world settings.
\end{abstract}

\section{Introduction}
\setstretch{0.96}
Sequential manipulation planning is an essential capability for robots to autonomously perform diverse tasks in complex environments. Executable motions must be effectively generated for robots to achieve long-term task objectives, requiring efficient planning algorithms for responsive operation and reasoning capabilities to anticipate environmental changes. \acf{tamp} formulates a methodology that hierarchically decomposes planning into two stages: the high-level symbolic task planning stage reasons over long-horizon abstract action sequences, and the low-level continuous motion planning stage computes feasible trajectories subject to geometric constraints. In recent years, \ac{tamp} has enabled significant advances~\cite{dantam2016incremental,toussaint2018differentiable,garrett2021integrated,jiao2021consolidated,jiao2021efficient,jiao2022sequential,su2023sequential}. However, a persistent challenge remains to properly interface between the task planner and the motion planner to efficiently solve \ac{tamp}, \ie, generating action sequences that satisfy both symbolic task goals and continuous motion constraints.

\begin{figure}[t!]
    \centering
    \includegraphics[width=\linewidth]{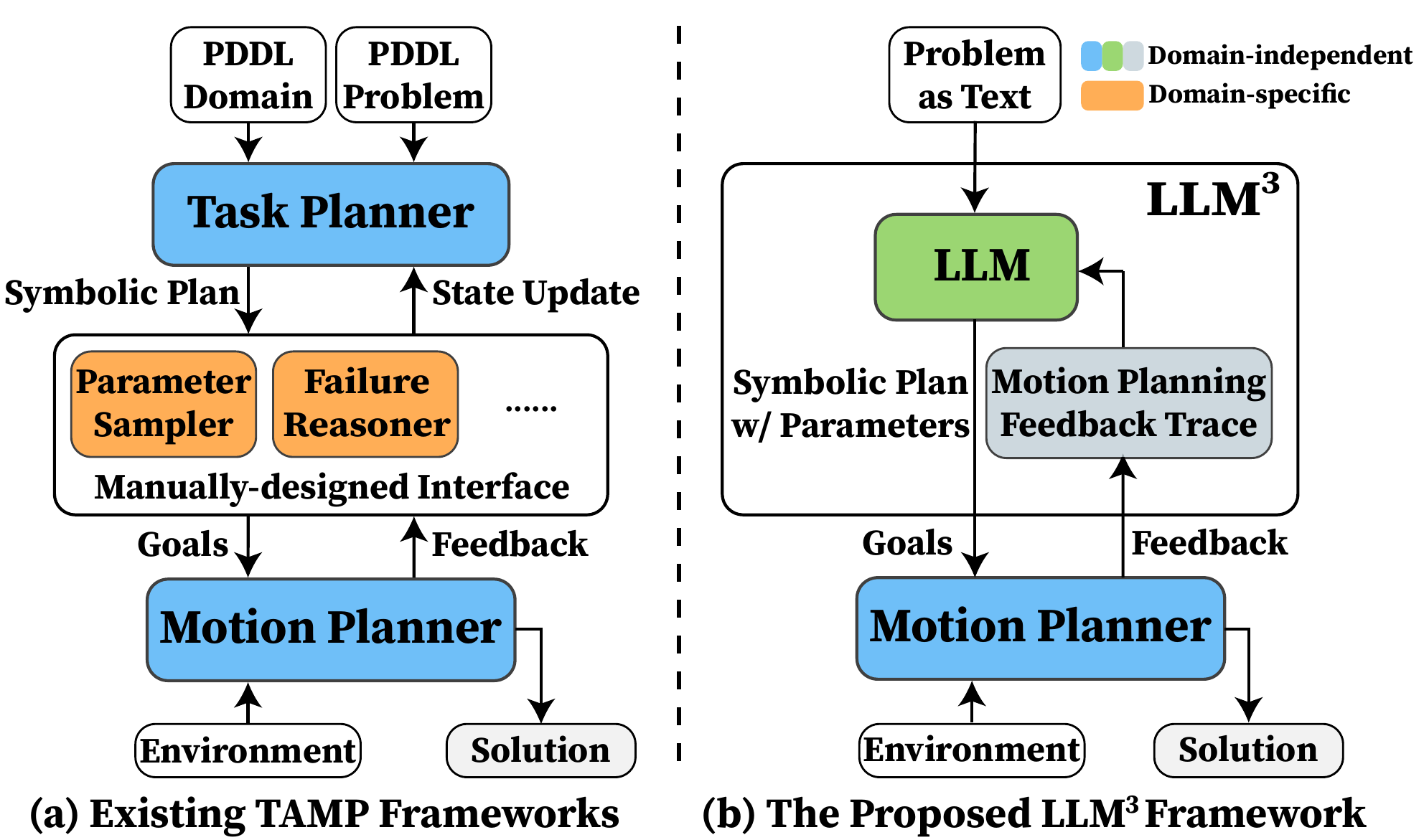}
    \caption{\textbf{The proposed \llmmm framework.} (a) Traditional \ac{tamp} frameworks rely on manually designed, domain-specific modules for interfacing between task and motion planners. (b) In contrast, we leverage a pre-trained \ac{llm} to iteratively propose refined plans and action parameters, by reasoning on motion planning failures. }
    \label{fig:teaser}
\end{figure}

Traditional \ac{tamp} approaches often rely on manually designed modules to interface between symbolic and continuous domains, as depicted in \cref{fig:teaser}(a). These modules serve two key roles. First, they act as action parameter samplers that generate real-valued parameters for symbolic actions. These parameters provide numerical goals to the motion planner. Selecting appropriate action parameters, \eg, in the object rearrangement task, a suitable 2D target location $(x, y)$ for action \texttt{Place(object)}, is crucial for the motion feasibility of actions and the efficiency of \ac{tamp}. While previous works propose to learn heuristic parameter samplers from data~\cite{chitnis2016guided,wang2018active}, they are tailored to specific domains and lack generalizability. Second, these modules implement mechanisms to incorporate motion failure into the task planner to generate improved action plans, \eg, by updating the symbolic state~\cite{srivastava2014combined}. However, they usually require domain-specific design by human experts. In summary, these modules are domain-specific and require substantial manual effort to design, which hinders generalizability to novel environments.

Recent \acfp{llm} pre-trained on web-scale text data have demonstrated emergent capabilities in reasoning~\cite{kojima2022large} and planning~\cite{huang2022language}. Pre-trained \acp{llm} can perform task planning~\cite{huang2023inner}, generate continuous parameters~\cite{mirchandani2023large}, and reason on environment feedback~\cite{huang2023inner}. Our intuition is that pre-trained \acp{llm} could provide a general and domain-independent approach to interfacing between symbolic and continuous domains for \ac{tamp}, eliminating the need to design domain-specific modules manually.

In this paper, we present \textbf{\llmmm} (\textbf{L}arge \textbf{L}anguage \textbf{M}odel-based Task and \textbf{M}otion Planning with \textbf{M}otion Failure Reasoning), an \ac{llm}-powered \ac{tamp} framework that reasons over motion planning feedback for effective planning\cref{fig:teaser}(b). Specifically, \llmmm employs a pre-trained \ac{llm} to (i) propose symbolic action sequences towards the task goal, (ii) generate continuous action parameters that lead to feasible motion, and (iii) reason over motion planning feedback to iteratively refine the proposed symbolic actions and parameters. This framework offers several key advantages over traditional \ac{tamp} approaches. First, it does not require manually designed symbolic domain files for task planning, instead leveraging the knowledge encoded in the \ac{llm} to propose symbolic actions. Second, it uses the \ac{llm} as a domain-independent informed parameter sampler to generate continuous action parameters, which benefits from the implicit heuristics of the \ac{llm}~\cite{mirchandani2023large}. Third, its reasoning over motion planning feedback is independent of the specific choice of planner. Crucially, we categorize and organize the possible motion planning feedback to feature two major motion failure modes, \ie, collision and unreachability. Such motion planning feedback allows \llmmm to refine the generated action sequences and parameters in a more targeted way, and find a feasible \ac{tamp} solution with fewer planning iterations and motion planning queries.

We evaluate \llmmm in a simulated tabletop box-packing task, which poses challenges in reasoning about potential failure modes, collisions, and unreachable areas, throughout the sequential manipulation planning problem. Quantitative results demonstrate the effectiveness of \llmmm, with ablation studies verifying: (i) reasoning over motion feedback significantly improves success rates and planning efficiency, and (ii) the \ac{llm}-based parameter sampler is substantially more sample efficient than a random sampler. Furthermore, we conduct real-robot experiments to show that \llmmm can be applied to real-world problems.

In summary, our contributions are threefold:
\begin{enumerate}
    \item We introduce \llmmm, the \textit{\textbf{first}} \ac{tamp} framework that holistically employs a pre-trained \ac{llm} as a domain-independent task planner, informed action parameter sampler, and motion failure reasoner.
    \item We categorize and organize the feedback from the motion planner, which enables \llmmm to efficiently identify and resolve planner-independent motion failures through targeted refinement of actions and parameters.
    \item We conduct comprehensive experiments in both the simulation and the real world to demonstrate the effectiveness of \llmmm in solving \ac{tamp} problems.
\end{enumerate}

\subsection{Related Work}

\paragraph*{Task and Motion Planning}
Conventional \ac{tamp} approaches employ a high-level task planner to generate symbolic action sequences and a low-level motion planner to generate motion trajectories. The task planner requires pre-designed symbolic planning domains represented in formatted representations, such as \ac{pddl}. Significant efforts have been made to develop manually engineered modules that interface the task planner and motion planner, such as incorporating motion-level constraints into task planning~\cite{garrett2020pddlstream,jiao2021efficient}, making approximations at the motion level~\cite{hauser2011randomized,toussaint2015logic}, and designing specialized communication modules~\cite{srivastava2014combined}. However, manually defining task planning domains and interface modules to fully capture real-world complexity is impractical. Furthermore, as the action space grows, searching for geometrically feasible symbolic action sequences becomes computationally challenging without effective heuristics~\cite{garrett2020pddlstream}. Recent work has explored data-driven heuristics to improve \ac{tamp} efficiency~\cite{chitnis2016guided,yang2023sequence}, but these domain-specific heuristics lack generalizability across domains. In this work, we employ a pre-trained \ac{llm} as both the task planner and the interface between task and motion. We expect that semantic knowledge in the \ac{llm} can provide domain-independent heuristics for \ac{tamp}.

\paragraph*{Robot Planning with \acp{llm}}
Recent \acfp{llm}~\cite{NEURIPS2023_ee6630dc, jiang2024raising} encode vast world knowledge and exibit the emergent capability for planning~\cite{huang2022language}. Pre-trained \acp{llm} have been applied for task planning of robots or embodied agents~\cite{huang2023inner,wang2023describe,wang2023voyager, han2024interpret, gong2023lemma, gong2023mindagent, ding2023quar, song2024germ}. Notably, Inner Monologue~\cite{huang2023inner} takes in textualized environment feedback and generate actions to execute, while ReAct~\cite{yao2022react} further advanced this closed-loop approach by integrating reasoning and acting. Voyager~\cite{wang2023voyager} and DEPS~\cite{wang2023describe} focus on developing open-ended embodied agents that iteratively replan based on execution failure. Furthermore, \acp{llm} have been applied to solve \ac{tamp} problems ~\cite{ding2023task,chen2023autotamp}. Specifically, LLM-GROP~\cite{ding2023task} employs a pre-trained \ac{llm} to instantiate symbolic goals and continuous object placements for semantic object rearrangement, which are then fed into a classical \ac{tamp} planner. AutoTAMP~\cite{chen2023autotamp} leverages an \ac{llm} to translate natural-language task specifications into a formal language processable by off-the-shelf \ac{tamp} algorithms. Our usage of \acp{llm} in \ac{tamp} is inspired by many of the above works; however, the major difference is that we leverage the \ac{llm} as the core component of our \ac{tamp} framework.



\begin{figure*}[t!]
    \centering
    \includegraphics[width=\linewidth,trim=0cm 0cm 0cm 0cm,clip]{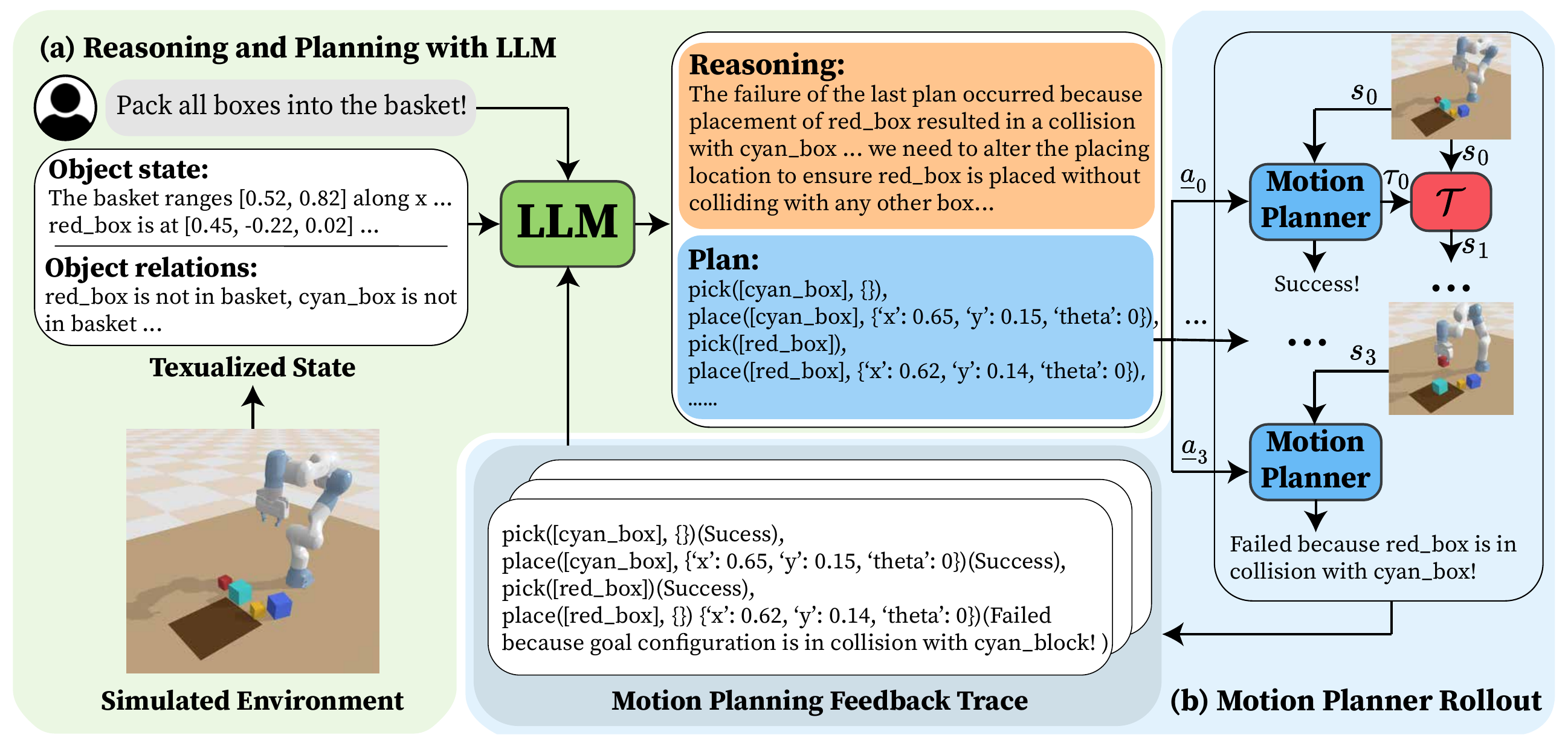}
    \caption{\textbf{System diagram of the proposed \llmmm framework.} (a) We show an example of utilizing a pre-trained \ac{llm} for reasoning and generating action sequences. (b) The feasibility of the proposed action sequence is verified by rollout with a motion planner and transition function $\mathcal{T}$. The motion planning feedback is saved into a trace that is provided to the \ac{llm} in the next iteration.}   
    \label{fig:framework}
\end{figure*}

\section{Preliminaries and Problem Setting}
\label{sec:prelim}

\subsection{Task and Motion Planning}

A \textit{\ac{tamp}} problem is a tuple $<\mathcal{O},\mathcal{S},\mathcal{A},\mathcal{T},s_0,g>$, where:
\begin{itemize}
    \item $\mathcal{O}$ is the set of objects in the environment.
    \item $\mathcal{S}$ is the state space factorized with respect to objects $\mathcal{O}$, where a state $s \in \mathcal{S}$ comprises the state of all objects, \eg, their 3D positions and dimensions.
    \item $\mathcal{A}$ is the set of primitive actions with low-level execution handled by a motion planner. A primitive action $a \in A$ is parameterized by object variables $\bar{o}$ and continuous parameters $\theta$, which can be instantiated with specific objects $\underline{o}=(o_1,...,o_m)$ and parameter values to produce a ground action $\underline{a}$, \eg, $\texttt{Place}(\texttt{block},[0.1,0.2])$. The parameters $\theta$ provide the goal for the motion planner; we say $\underline{a}$ is feasible when the motion planner has a solution, \ie, a collision-free trajectory $\tau$. We denote a feasible action by $\underline{a}(\tau)$.
    \item $\mathcal{T}$ is the state transition function that outputs the next state $s_{t+1}$ after executing an action $\underline{a}_t(\tau_t)$ at a state $s_t$, \ie, $s_{t+1}=\mathcal{T}(s_t,\underline{a}_t(\tau_t))$. We assume that $\mathcal{T}(s_t,\underline{a}_t(\tau_t)))$ can be evaluated with a black-box simulator.
    \item $s_0$ is the initial state that follows $s_0 \in \mathcal{S}$.
    \item $g$ is the goal function $g:~\mathcal{S} \rightarrow \{0,1\}$, which checks whether the task goal is achieved at state $s$. 
\end{itemize}
The objective of \ac{tamp} is to derive a sequence of feasible actions $(\underline{a}_0(\tau_0),\underline{a}_1(\tau_2),..., \underline{a}_T(\tau_T))$, such that $g(s_{T+1})=1$ and $s_{t+1}=\mathcal{T}(s_t,\underline{a}_t(\tau_t)))$, where $t=0,1,...,T$.


A common strategy to solve the \ac{tamp} problem is known as ``search-then-sample''~\cite{srivastava2014combined,dantam2016incremental,garrett2021integrated}, which alternates generating symbolic action sequences through backtracking search and sampling the continuous action parameters until a feasible plan is found to reach the goal. However, a naive search-then-sample task and motion planner is usually inefficient, as verifying the action feasibility requires invoking the computationally expensive motion planning process~\cite{cui2024anyskill, li2024ag2manip}. To mitigate this complexity, researchers have crafted symbolic domains~\cite{srivastava2014combined,silver2023predicate, qian2024learning} to prune infeasible symbolic action sequences, or learned heuristic samplers to sample action parameters~\cite{wang2018active,fang2023dimsam} that lead to feasible plans. In this work, we leverage pre-trained \acp{llm} as both a task planner and a heuristic sampler to generate symbolic action sequences and action parameters.

\subsection{Planning as Sequence Prediction}

Recently, planning problems have been formulated as sequence prediction to avoid the computationally expensive search process~\cite{driess2020deep}. In particular, pre-trained \acp{llm} are employed to generate discrete actions~\cite{huang2022language} and continuous parameters in an auto-regressive manner when provided in-context prompts. Formally, with textualized initial state $s_0$, goal $g$ and additional context $c$, we have:
\begin{equation}
    \small
    \begin{aligned}
        \underline{a}_{0:T} 
        &= \arg\max~p_{LM}(\underline{a}_{0:T}|s_0,g,c) \\
        &= \arg\max~p_{LM}(\underline{a}_0|,s_0,g,c) \prod_{t=1}^T p_{LM}(\underline{a}_t|\underline{a}_{0:t-1},s_0,g,c),
    \end{aligned}
\end{equation}
where $p_{LM}$ is the generative probability of a pre-trained language model. Following this scheme, we use pre-trained \acp{llm} to propose sequences of symbolic actions and continuous action parameters based on the initial state, goal, and trace of motion planning feedback. As the proposed action sequences may be infeasible at the motion level, we validate the feasibility of proposed action sequences with a motion planner, the state transition function $\mathcal{T}$ and the goal $g$. We iterate these two steps until a feasible plan is found.

\section{Method}
\label{sec:method}
We introduce \llmmm, a \ac{tamp} framework that leverages a pre-trained \ac{llm} to reason on motion failure and generate iteratively refined symbolic actions and continuous action parameters. The system diagram of \llmmm is shown in \cref{fig:framework}. Below, we elaborate on the overall framework, reasoning and planning with the pre-trained \ac{llm}, and the designed motion planning feedback.

\begin{algorithm}[t!]
\small
    \DontPrintSemicolon 
    \caption{\llmmm for \ac{tamp}}
    \label{alg:llm3}
    \LinesNumbered
    \SetKwInOut{KIN}{Input}
    \SetKwInOut{KOUT}{Output}
    
    \KIN{pre-trained LLM with prompt template to generate action sequences $LLM$, state transition function $\mathcal{T}$ with motion planner $MP$, goal function $g$, initial state $s_0$, maximum number of planning attempts $N_{max}$, maximum trace size $k$}
    \KOUT{success indicator $success$, a sequence of feasible actions $plan$}

    \nonl \textcolor{blue}{// initialize output and feedback trace} \;
    $plan \gets []$, $success \gets \text{False}$, $feedback \gets []$, $trace \gets []$ \;
    \nonl \textcolor{blue}{// initialize current state and planning iteration count} \;
    $s \gets s_0$, $iter \gets 0$ \;
    \nonl \textcolor{blue}{// main planning loop} \;
    \While{not $success$ and $iter < N_{max}$}{
        \nonl \textcolor{blue}{// reason and plan with LLM }\;
        $reason,llm\_plan \gets LLM(s_0,trace)$, $iter \gets iter + 1$\; 
        \nonl \textcolor{blue}{// rollout $llm\_plan$ from $s$ with motion planner}\;
        \ForEach{$\underline{a} \in llm\_plan$}{
            \nonl \textcolor{blue}{// call motion planner to verify feasibility of \underline{a}}\;
            $\tau, mp\_feedback \gets MP(s;\underline{a})$\;
            $feedback.append((\underline{a},mp\_feedback))$\;
            \nonl \textcolor{blue}{// terminate rollout if action $\underline{a}$ is infeasible}\;
            \If{$\tau$ is $None$}{
                \textbf{break}\;
            }
            \nonl \textcolor{blue}{// update state $s$ and add $\underline{a}$ to $plan$}\;
            $s \gets \mathcal{T}(s,\underline{a}(\tau))$, $plan.append(\underline{a}(\tau))$\;
        }
        \nonl \textcolor{blue}{// check whether $s$ satisfies goal}\;
        $success, task\_feedback \gets g(s)$\;
        \nonl \textcolor{blue}{// when the plan is feasible but doesn't reach goal}\;
        \If {not $success$}{
            \nonl \textcolor{blue}{// record $task\_feedback$}\;
            $feedback.append(task\_feedback)$\;
            \nonl \textcolor{blue}{// add aggregated feedback to $trace$}\;
            $trace.append(feedback)$, $trace \gets trace[-k:]$\;
            \nonl \textcolor{blue}{// clear aggregated feedback}\;
            $feedback \gets []$\;
        }
    }
    \Return{$success$, $plan$}\;
\end{algorithm}

\subsection{The \texorpdfstring{\llmmm}{} Framework}
\label{subsec:overall}

As shown in Algorithm \cref{alg:llm3} and Figure \ref{fig:framework}, the \llmmm framework iterates between: (i) reasoning on previous motion failure and generating an action sequence (\ie, symbolic actions and continuous parameters) with a pre-trained \ac{llm}, and (ii) verifying the feasibility of the action sequence with a motion planner. At each planning iteration, the \ac{llm} takes the current state $s$ and the trace of motion planning feedback $trace$, and outputs the reasoning for the previous motion failure $reason$ and an action sequence $llm\_plan$ to solve the \ac{tamp} problem (see \cref{subsec:llm_tamp}). The motion planner then attempts to find a collision-free motion trajectory for each action $\underline{a} \in llm\_plan$ sequentially. We synthesize motion planning feedback (see \cref{subsec:mp_fb}) for each motion planner query and aggregated the feedback for all actions in $llm\_plan$. The planning iteration ends with failure when an action has no feasible motion or the action sequence fails to reach the goal. The aggregated feedback is then added to a trace $trace$, which is maintained throughout the life cycle of the framework with a maximum size $k$. In the next planning iteration, the trace is fed into the \ac{llm} to generate motion failure reasoning and another action sequence that improves on the previous one. This process repeats until a generated action sequence reaches the goal with no motion failure or the maximum number of attempts is exceeded. Overall, the \llmmm framework can be regarded as a search-then-sample \ac{tamp} planner that generates action sequences with incrementally improved quality, guided by the intrinsic heuristics of the pre-trained \ac{llm} and the previous motion failure. By design, we expect \llmmm to exhibit superior efficiency compared to unguided planners that sample action parameters randomly.

\subsection{Reasoning and Planning with pre-trained LLM}
\label{subsec:llm_tamp}

Following previous attempts in utilizing \acp{llm} for reasoning and planning~\cite{kojima2022large,huang2022language,huang2023inner}, we prompt a pre-trained \ac{llm} to generate motion failure reasoning and action sequences in text format. Since we want to limit the domain-specific prior provided to the \ac{llm}, we use zero-shot prompting without providing any planning examples. We adopt \ac{cot} prompting to have the \ac{llm} generate the reasoning and action sequence in an auto-regressive manner. The prompt fed into \ac{llm} comprises the following contents: (i) a system message that provides the global context to the pre-trained \ac{llm} and activates its planning capability, (ii) a task description specifying the environment, goal, and available primitive actions, (iii) the textualized initial environment state, (iv) the trace of motion planning feedback, and (v) the output format. Note that most promp content will remain unchanged for different planning iterations of the same \ac{tamp} problem. Only the motion planning feedback trace in (iv) will be updated as new failed action sequences are added. \cref{fig:framework}(a) illustrated an example of reasoning and planning with \ac{llm}. 

We implement two strategies for the pre-trained \ac{llm} to generate a new action sequence that improves on the previous one: (i) \textit{backtrack}, where we expect the \ac{llm} to backtrack to a previous action that has feasible motion, and continual to generate actions that complete the plan, and (ii) \textit{from scratch}, where we expect the \ac{llm} to directly generate a new action sequence that attempts to avoid the motion failure happened to its previous output. We achieve this by designing the system message and output format description in the prompt. We show the prompt template for the two variants in \cref{fig:prompt}, where the content specific to each variant is highlighted.


\begin{figure}[ht!]
    \centering 
    \framedtext{
    \textrm{\small 
    \vspace{-1mm} \\
    \textbf{System message}: You are an AI robot that generates a plan of actions to reach the goal... \hllo{You are expected to correct the plan incrementally (on top of the last plan) to avoid motion failure. This may involve sample new parameters for the failed action or reverse one or more succeeded actions for backtracking...} \hllb{You are expected to generate a plan from scratch.}\\
    \textbf{Task description:} A robot arm is tasked to pack boxes into a basket on a table. The robot sits at (0, 0), and faces the positive x-axis, while the positive z-axis points up.
    The robot is equipped with primitive actions, each taking a list of objects and continuous parameters as input:
    \begin{itemize}
        \item $\texttt{pick}([\texttt{obj}], \{\})$: pick up \texttt{obj}, with no parameters.
        \item $\texttt{place}([\texttt{obj}], \{``x``: [0.0, 1.0], ``y``: [-1.0, 1.0], \\``theta``: [-3.14, 3.14]\})$: place \texttt{obj} at location $(x, y)$ with the planar rotation theta, where $x$ ranges $(0.0, 1.0)$, $y$ ranges $(-1.0, 1.0)$, and $theta$ ranges $(-3.14, 3.14)$.
    \end{itemize}
    \textbf{Initial state}: \{init\_state\}\\
    \textbf{motion planning feedback trace}: \{trace\}\\
    \textbf{Output format:} Please generate output step-by-step, which includes your reasoning for the failure of the last plan as well as the generated plan... Answer the questions: (i) what is the cause of the failure of the last plan? \hllo{(ii) can altering action parameters for the failed action solve the problem... (iii) do we need to reverse one or more succeeded actions executed before the failed action...} \hllb{(ii) what is your strategy to generate a new plan from scratch to accomplish the task goal?}
    Please organize the output following the JSON format below:
    \{
        ``Reasoning'': ``My reasoning for the failure of the last plan is ...'',
        ``Full Plan'': [``pick(['red\_box'], \{\})'', ``place(['red\_box'], \{'x': 0.51, 'y': 0.02, 'theta': 0.00\})'', ...]
    \}
    }
    }
    \vspace{-2mm}
    \caption{\textbf{Prompt templetes used by \llmmm.} We show alternative contents specific for the \textit{backtrack} variant in \hllo{orange} and \textit{from scratch} variant in \hllb{blue}. } 
    \label{fig:prompt}
\end{figure}

\subsection{Synthesizing Motion Planning Feedback}
\label{subsec:mp_fb}

We realize a ground action $\underline{a}$ by calculating a collision-free trajectory $\tau$ with a sampling-based motion planner, \eg Bi-directional Rapidly-exploring Random Trees (BiRRT)~\cite{lavalle2006planning}. The input to the motion planner includes the initial environment state (includes the robot state), and a goal pose of the robot end effector specified by the continuous action parameters of $\underline{a}$. The motion planner samples and searches for a collision-free joint space trajectory for the robot to reach the goal end-effector pose.

By default, the motion planner reports a binary signal that indicates whether there is a feasible trajectory. It does not give more abstract-level feedback, which explains why motion planning fails. As a result, the \ac{tamp} planners can acquire useful feedback from motion planning failures to improve high-level planning. To this end, we additionally synthesize semantically meaningful motion-level feedback so that \llmmm can improve on previous failures more effectively. We observe that typical motion planning failures can be categorized into two types, \ie, collisions and unreachability. Therefore, we synthesize categorized motion planning feedback following the templates below (\cref{fig:feedback}):

\begin{figure}[!b]
    \centering
    \includegraphics[width=\linewidth,trim=0cm 0cm 0cm 0cm,clip]{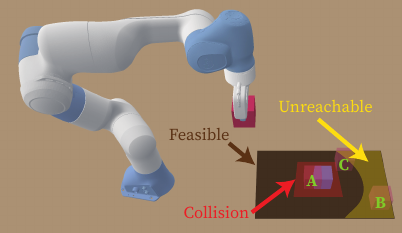}
    \caption{\textbf{Three types of motion possibilities.} A: the object placement is in collision with an existing object. B: the object placement is beyond the robot's reach. C: the object placement is feasible. }   
    \label{fig:feedback}
\end{figure}

\begin{enumerate}[label=(\Alph*)]
    \item The goal configuration is in collision with \texttt{object}.
    \item The goal configuration has no feasible IK solution.
    \item The goal configuration is collision-free and reachable.
\end{enumerate}
In practice, we integrate the motion planner with an additional IK solver and collision checker for obtaining these feedbacks, finding this design to be practically effective.

\section{Simulation and Experiment}\label{sec:sim}

In simulations, we initially perform an ablation study on our \llmmm framework in two settings of the tabletop box-packing task, quantitatively evaluating its effectiveness based on i) the planning success rate (\%SR), ii) the number of \ac{llm} calls (\#LM), iii) and the number of motion planner calls (\#MP). Additionally, we demonstrate the role of \ac{llm} as an informed action parameter sampler by comparing it to a baseline utilizing random sampling strategies. Finally, we validate the proposed \llmmm framework through experimentation on a perception-integrated physical robotic manipulator, confirming its validity in real-world scenarios.

\subsection{Simulation Setup}
\label{subsec:sim_setup}
\begin{figure}[t!]
    \centering
    \includegraphics[width=\linewidth,trim=0cm 0cm 0cm 0cm,clip]{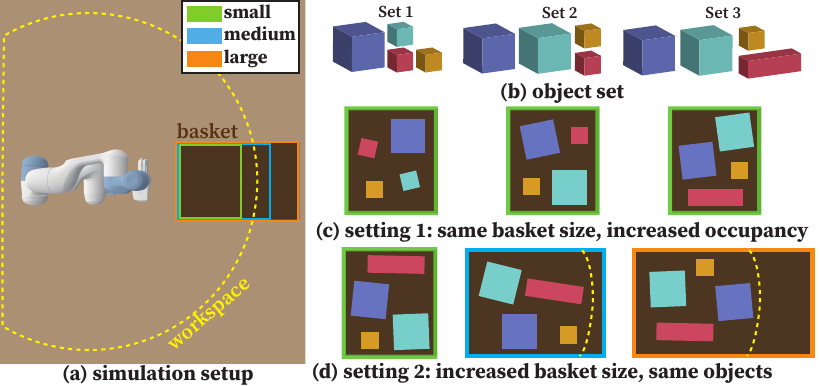}
    \caption{\textbf{The box-packing task setup in a simulated environment.} (a) The task requires the robot to place one of (b) three sets of objects fully into the basket. (c) In setting 1, the total object size increases but the basket sizes remain the same. All baskets are fully reachable by the robot. (d) In setting 2, the basket size increases, but some portions of baskets are longer within the robot's reach.} 
    \label{fig:sim_setup}
\end{figure}

The simulation has an important role in Robotics\cite{gao2019vrkitchen, gao2020joint}. We developed a PyBullet-based simulation environment for our box-packing tasks, as illustrated in \cref{fig:sim_setup}a. In \textbf{Setting 1}, three different sets of objects are given with increasing total sizes, while the basket size remains unchanged. This task requires the \llmmm to oversee potential collisions among objects and the robot throughout the action sequence. The \ac{llm} must reason why collisions occur and adjust previous actions to ensure feasible task and motion plans. \textbf{Setting 2} involves placing the Set 3 objects into baskets of increasing sizes. Here, the robot cannot access the entire basket region but encounters a collision likelihood similar to the most crowded condition in Setting 1. In this setting, the task becomes more complex as it challenges the \ac{llm} to reason about the robot's operational space and adjust the plan accordingly. Throughout the simulations, we utilize GPT-4 Turbo as the \ac{llm} planner and BiRRT~\cite{lavalle2006planning} as the motion planner, with 10 attempts for each setting. \footnote{The code is available at \href{https://github.com/AssassinWS/LLM-TAMP}{https://github.com/AssassinWS/LLM-TAMP}.}

\subsection{Ablation Study}
The conducted ablation study compares the proposed \llmmm with baseline methods:

\begin{table*}[!t]
    \centering
    \caption{Ablation Study}
    \resizebox{\linewidth}{!}{%
        \begin{tabular}{ccccccccccccccccccc}%
            \toprule
            \multirow{3}{*}{\textbf{Method}} &
            \multicolumn{10}{c}{\textbf{Setting 1}} &
            \multicolumn{8}{c}{\textbf{Setting 2}} \\
            \cmidrule(lr){2-10}\cmidrule(lr){11-19} &
            \multicolumn{3}{c}{\textbf{Easy}} &
            \multicolumn{3}{c}{\textbf{Medium}} &
            \multicolumn{3}{c}{\textbf{Hard}}  &
            \multicolumn{3}{c}{\textbf{Small}} &
            \multicolumn{3}{c}{\textbf{Medium}} &
            \multicolumn{3}{c}{\textbf{Large}}\\
            \cmidrule(lr){2-4}\cmidrule(lr){5-7}\cmidrule(lr){8-10}
            \cmidrule(lr){11-13}\cmidrule(lr){14-16}\cmidrule(lr){17-19} & 
            \textbf{\%SR} & \textbf{\#LM} & \textbf{\#MP}& 
            \textbf{\%SR} & \textbf{\#LM} & \textbf{\#MP}&
            \textbf{\%SR} & \textbf{\#LM} & \textbf{\#MP}&
            \textbf{\%SR} & \textbf{\#LM} & \textbf{\#MP}&
            \textbf{\%SR} & \textbf{\#LM} & \textbf{\#MP}&
            \textbf{\%SR} & \textbf{\#LM} & \textbf{\#MP}\\
            
            \midrule
            \llmmm Backtrack& \textbf{100} & \textbf{1.6} & \textbf{11.8} & \textbf{100} & \textbf{4.4} & \textbf{28.4} & 60 & 11.4 & 39.8 & 60 & 11.4 & 39.8 & \textbf{80} & \textbf{9.5} & \textbf{50} & 50 & 13.5 & 44.8 \\
            Backtrack & 100 & 1.8 & 12.6 & 90 & 6.3 & 32 & 40 & 15.1 & 55.3 & 40 & 15.1 & 55.3 & 30 & 14.6 & 16 & 30 & 15.8 & 48  \\
            \llmmm Scratch & 100 & 1.7 & 13.2 & 100 & 7 & 46.1 & \textbf{70} & \textbf{8.8} & \textbf{30.9} & \textbf{70} & \textbf{8.8} & \textbf{30.9} & 70 & 11.5 & 50.2 & \textbf{60} & \textbf{10.6} & \textbf{32} \\
            Scratch & 100 & 2.4 & 17.6 & 60 & 12.3 & 45.3 & 50 & 13.7 & 42.8 & 50 & 13.7 & 42.8 & 30 & 16.2 & 45.7 & 40 & 13.2 & 24  \\

            \bottomrule
        \end{tabular}
    }
    \label{table1}
\end{table*}

\begin{enumerate}
    \item \textbf{\llmmm Backtrack}: The proposed \llmmm framework \textit{backtrack} variant. See \cref{fig:prompt}
    \item \textbf{Backtrack}: The \ac{llm} proposes plans with backtracking but without motion planning feedback (line 7).
    \item \textbf{\llmmm Scratch}: The proposed \llmmm framework \textit{from scratch} variant. It replans the entire action sequence, incorporating feedback from the motion planner if any action fails. See \cref{fig:prompt}
    \item \textbf{Scratch}: The \ac{llm} plans the action sequence once and executes the plan without any feedback.
\end{enumerate}

Three evaluation criteria are considered: The number of \ac{llm} calls (\#LM) counts how many times the \ac{llm} API is called during planning. A lower \#LM indicates that the planner can produce a feasible task plan more efficiently. In each attempt, \#LM has a maximum cap of 20; attempts with over 20 \#LMs will be counted as a failure. The total success rate is recorded as \%SR. Additionally, the number of motion planner calls, \#MP, is another critical criterion as in traditional \ac{tamp} approaches, where massive and time-consuming motion planner calls are the main cause for their inefficiency. The study results are summarized in \cref{table1}.

In both settings, integrating motion planning feedback results in a decrease in \#LM and \#MP, along with an increase in \%SR for both \textit{backtrack} and \textit{scratch} strategies. This indicates that the \ac{llm} can reason about failures from motion planning feedback, and importantly, propose adjusted task plans and action parameters that are more likely to produce feasible motions. Surprisingly, no clear evidence suggests that utilizing backtracking is superior to replanning from scratch. We've observed distinct behaviors among different strategies employed by \llmmm. Specifically, when utilizing backtracking, \ac{llm} sometimes consistently adjusts the action parameter of the specific action that \textit{directly} led to failure, without adjusting previous actions. In contrast, the \llmmm employing a planning from scratch strategy simply re-samples all actions and parameters, occasionally resulting in a slightly lower \#LM.

\subsection{Action Parameter Selection}
\begin{table}[!b]
    \vspace{5pt}
    \small
    \centering
    \caption{Comparison of Different Sampling Strategies}
        \begin{tabular}{cccc}
            \toprule
            \textbf{Method} & \textbf{\#Itration} & \textbf{\#MP}  \\
            \midrule
             Random Samples & 109.6 & 663.1\\
             LLM &  10.8 & 70.2 \\
             LLM + Feedback & \textbf{7.9} & \textbf{53.2} \\ 
            \bottomrule
        \end{tabular}
    
    \label{table2}
\end{table}

This study examines whether an \ac{llm} can function as an informed action parameter sampler, using the \textbf{Medium} setup from \textbf{Setting 1} in the previous section. In this study, the action sequence is predetermined and consists of a total of 8 steps, where the robot sequentially performs pick and place actions to relocate all four blocks into the basket. However, the specific placement location is not provided, requiring the action parameters to be sampled to ensure the feasibility at the motion level. We implement three methods, each runs 50 attempts, for comparison:

\begin{enumerate}
    \item \textbf{Random Samples}: A random sampler is implemented to independently and uniformly sample all the block placement locations within the basket region.
    \item \textbf{\ac{llm}}: The \ac{llm} is provided with the task setting and the symbolic action sequence, prompting it to select the action parameters to make the action sequence feasible.
    \item \textbf{\ac{llm} + Feedback}: Building upon 2), the \ac{llm} is additionally provided with motion planning feedback if the previous attempt fails.
\end{enumerate}


The results summarized in \cref{table2} demonstrate the efficacy of using \ac{llm} as an informed action parameter sampler in the context of box-packing tasks. Specifically, while random sampling requires an average of 109.6 iterations and 663.1 \#MP to achieve feasible action sequences, the \ac{llm} substantially reduces them to an average of 10.8 iterations and 70.2 \#MP. When incorporating motion planning feedback alongside the \ac{llm}, the sampling requirement decreases even further to an average of 7.9 iterations and 53.2 \#MP. These findings underscore the ability of \acp{llm} to efficiently select action parameters that align with task constraints, resulting in a notable reduction in the number of motion planning calls needed to generate feasible action sequences. Moreover, the additional benefit gained from integrating motion planning feedback highlights the importance of incorporating real-time feedback mechanisms to further refine and optimize the sampling process. Overall, these results highlight the potential of \ac{llm}s as a valuable tool to improve the efficiency and effectiveness of robotic manipulation tasks, particularly in scenarios where efficient action parameter selection is crucial for improving planning performance.

\subsection{Experiment Setup}\label{sec:exp}
\begin{figure*}[t!]
    \centering
    \includegraphics[width=\linewidth,trim=0cm 0cm 0cm 0cm,clip]{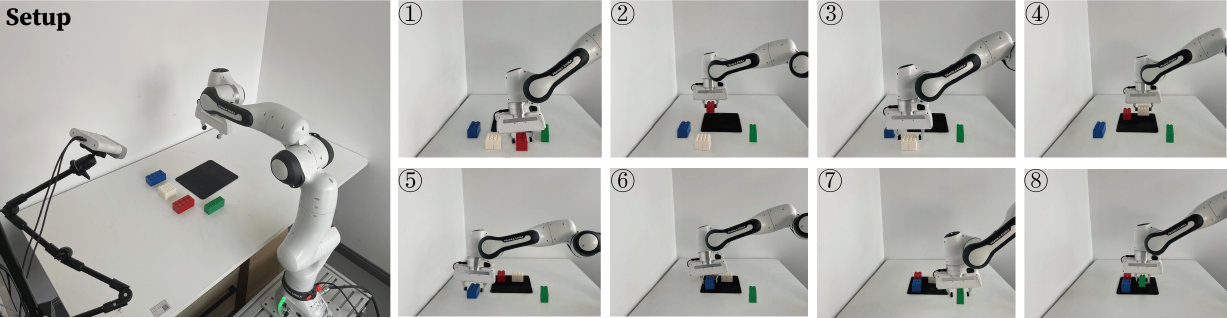}
    \caption{\textbf{The real-world experiment on a physical robot.} The figure to the left shows the box-packing task setup. Actions \protect\circled{1} to \protect\circled{8} are proposed by \llmmm and successfully carried out by the physical manipulator.}   
    \label{fig:exp_setup}
\end{figure*}

To validate the effectiveness of our proposed method in a real-world setting, we conducted an experiment using a Franka~Research~3 manipulator. The goal was to demonstrate the robot's ability to perceive and manipulate objects in an environment with uncertainties in both perception and execution. The robot observed a single point cloud from a third-person-view RGB-D camera, capturing the workspace containing various objects such as blocks and a plate. To identify and locate individual objects, we employed Grounded Segment Anything~\cite{ren2024grounded} for object segmentation, initially segmenting objects in the 2D RGB image and then projecting the results onto the corresponding 3D point cloud. This approach yielded per-object point clouds, essential for planning and executing manipulation tasks.

\cref{fig:exp_setup} presents a qualitative evaluation of our method, where the robot was tasked with placing all blocks on the plate. The results demonstrate that our method enabled the robot to successfully identify and manipulate objects despite the uncertainties and challenges in the cluttered environment. The successful execution of this experiment validates the practicality and robustness of our approach, showcasing its potential for various real-world applications such as object sorting, assembly tasks, and household assistance.

\section{Conclusions} \label{sec:conclude}
In this paper, we introduced \llmmm, a new \ac{tamp} framework powered by the pre-trained \ac{llm}. \llmmm leverages the rich knowledge encoded in and the powerful reasoning capability processed by \acp{llm} to (i) propose action sequences based without requiring a prior planning domain, (ii) generate continuous action parameters for the robot motion planner, and more importantly, (iii) refine task and/or motion plans in response to motion planning failures.

Following validation through various simulations and experiments, we demonstrated that \llmmm effectively produced and refined task and motion plans for box-packing problems, exhibiting promising potential in addressing previously unspecified tasks. Our study also revealed that although the pre-trained \ac{llm} can generate action parameters more efficiently than random samplers, it still necessitated multiple feedback iterations and motion planner calls. Looking ahead, the incorporation of in-context learning or fine-tuning techniques holds promise for further enhancing its efficiency, representing a crucial step towards empowering robots to tackle emerging tasks in real-world scenarios.

\setstretch{0.9}
{
\tiny
\balance
\bibliographystyle{ieeetr}
\bibliography{IEEEfull}
}

\end{document}